%% file: paper.tex
\documentclass[review]{elsarticle}

\makeatletter
\def\@author#1{\g@addto@macro\elsauthors{\normalsize%
    \def\baselinestretch{1}%
    \upshape\authorsep#1\unskip\textsuperscript{%
      \ifx\@fnmark\@empty\else\unskip\sep\@fnmark\let\sep=,\fi
      \ifx\@corref\@empty\else\unskip\sep\@corref\let\sep=,\fi
    }%
    \def\authorsep{\unskip,\space}%
    \global\let\@fnmark\@empty
    \global\let\@corref\@empty  
  \global\let\sep\@empty}%
  \@eadauthor={#1}
}
\makeatother

\def\ack{\section*{Acknowledgements}%
  \addtocontents{toc}{\protect\vspace{6pt}}%
  \addcontentsline{toc}{section}{Acknowledgements}%
}

\usepackage{lineno,hyperref}
\usepackage{amsmath,epsfig}
\usepackage{dsfont}
\usepackage{color}
\usepackage{graphicx}
\usepackage{paralist}
\usepackage{caption}
\usepackage{subcaption}
\usepackage{ulem}
\usepackage[T1]{fontenc}
\usepackage{blindtext}
\usepackage[english]{babel}
\modulolinenumbers[1]
\usepackage{algorithm}
\usepackage{algorithmic}

\biboptions{sort&compress}
\def\mathbi#1{\textbf{\textit{#1}}}

\newcommand{\argmax}{\operatornamewithlimits{argmax}}
\newcommand{\figref}[1]{Fig.~\ref{#1}}
\newcommand{\secref}[1]{Section~\ref{#1}}
\newcommand{\equref}[1]{Eq.~(\ref{#1})}
\newcommand{\algref}[1]{Algorithm~\ref{#1}}

\newcommand{\etal}{\textit{et al.}}

\newcommand{\ie}{e.g., }

\newcommand{\sectionref}[1]{Section~\ref{#1}}
\newcommand{\equationref}[1]{Eq.~(\ref{#1})}
\newcommand{\tableref}[1]{Tab.~\ref{#1}}

\begin{document}

\begin{frontmatter}

\title{Abrupt Motion Tracking via Nearest Neighbor Field Driven Stochastic Sampling}


\author[BIT]{Tianfei Zhou}
\author[BIT]{Yao Lu\corref{cor1}}
\ead{vis\_yl@bit.edu.cn}
\cortext[cor1]{Corresponding author. Present address: 5 South Zhongguancun Street, Haidian District, Beijing 100081, China.}

\author[BIT]{Feng Lv}
\author[BIT]{Huijun Di}
\author[BIT]{Qingjie Zhao}
\author[UTS]{Jian Zhang}

\address[BIT]{Beijing Laboratory of Intelligent Information Technology, School of Computer Science, Beijing Institute of Technology, Beijing, China}
\address[UTS]{Advanced Analytics Institute, University of Technology, Sydney, Australia}

\begin{abstract}
  \input{abstract}

\end{abstract}


\end{frontmatter}


\input{introduction}
\input{related_work}

\input{annf}
\input{ssamc}

\input{experiments}
\input{conclusion}

\ack{
We are thankful for the anonymous reviewers for their suggestions helping us to improve this work.
We also acknowledge the support of
the National Natural Science Foundation of China (No.
61273273) and by Research Fund for the Doctoral Program
of Higher Education of China (No. 20121101110034).
}

\section*{References}

\bibliography{paper}

\end{document}

%% file: abstract.tex
Stochastic sampling based trackers have shown good performance for abrupt motion tracking
so that they have gained popularity in recent years.
However, conventional methods tend to use a two-stage sampling paradigm,
in which the search space needs to be uniformly explored with an inefficient preliminary sampling phase.
In this paper, we propose a novel sampling-based method
 in the Bayesian filtering framework to address the problem.
Within the framework,
nearest neighbor field estimation is utilized to compute the importance proposal probabilities,
which guide the Markov chain search towards promising regions
and thus enhance the sampling efficiency;
given the motion priors, a smoothing stochastic sampling Monte Carlo algorithm
is proposed to approximate the posterior distribution through a smoothing weight-updating scheme.
Moreover, to track the abrupt and the smooth motions simultaneously,
we develop an abrupt-motion detection scheme which can discover the presence of abrupt motions during online tracking.
Extensive experiments on challenging image sequences demonstrate the effectiveness and the robustness of our algorithm
in handling the abrupt motions.

%% file: keywords.tex
Visual tracking \sep abrupt motion \sep stochastic sampling \sep nearest neighbor field \sep Markov Chain Monte Carlo

%% file: introduction.tex
\section{Introduction}\label{section:intro}

Visual tracking can be viewed as a process of establishing temporal coherent relations between consecutive frames.
Applications of visual tracking have been commonly found in surveillance \cite{stauffer2000learning, benfold2011stable},
 human-computer interaction \cite{kim2008face} and medical imaging \cite{paragios2003level}, etc.
Although great performance improvement has been achieved so far,
the problem is still very challenging, especially in real-world scenarios that usually contain abrupt motions.
Most existing approaches provide inferior performance when encountered abrupt motions
because of their susceptibility to the motion discontinuity.
In this work, we seek to develop an effective sampling-based algorithm to address the abrupt motion tracking problem.

\begin{figure}[t]
  \centering
  \includegraphics[width=0.8\textwidth]{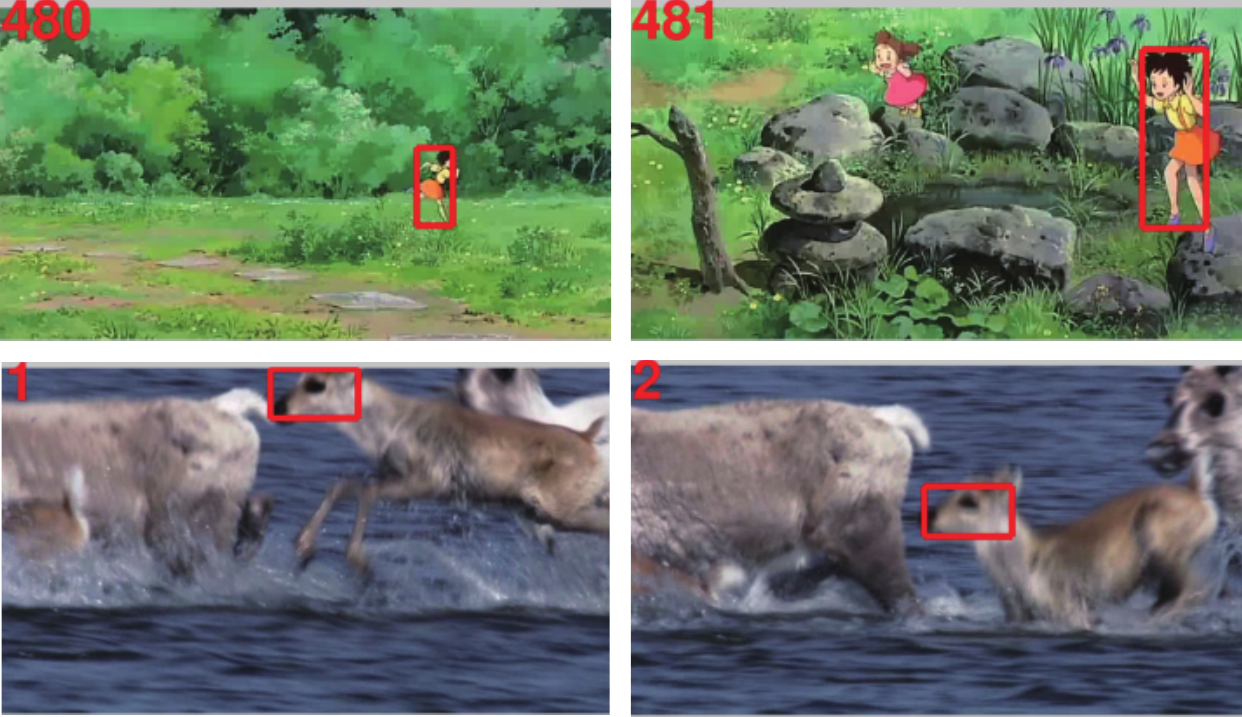}
  \caption{Examples of two abrupt motion scenarios. Top Row: \textit{shot change}. Bottom Row: \textit{Fast Motion}.
    Our tracking method successfully tracks the targets in these situations.
  }
  \label{fig:intro}
\end{figure}

Here, \textit{abrupt motion} is defined as the sudden changes of an object's location.
It may occur with various reasons: fast motion, shot changes, and low-frame-rate data source, etc.
\figref{fig:intro} illustrates some examples of the first two situations.
Developing a robust tracking algorithm in such complex scenarios is rather challenging,
 and several problems need to be thoroughly resolved:

First, most existing approaches cannot capture the unexpected object dynamic.
Particle filter(PF) has been demonstrated as a powerful method
to deal with the non-Gaussian and the multi-modal state space for visual tracking
(\ie \cite{perez2002color,nummiaro2003adaptive,LSST,CoupleLayer,zhong2014robust,Li2007cascade,jia2012visual, dou2014robust}).
Although performing well in low-dimensional systems, these methods have to draw a large number of particles
to guarantee sufficient sampling when abrupt motion occurs in which case the posterior density is very complex.
The large computational cost makes PF infeasible for practical applications.
Recently, Markov Chain Monte Carlo(MCMC) \cite{gilks2005markov, MCMC-overview} is widely used as an effective alternative of PF
because of the high computational efficiency in high-dimensional sample space.
However, it has been shown that MCMC-based tracking methods \cite{khan2004mcmc, khan2005mcmc} are prone to getting trapped in local modes
when the energy landscape of the posterior distribution is rugged.

The second problem that has not been addressed by previous studies
is tracking the abrupt and the smooth motions at the same time.
In abrupt motion tracking, it is commonly assumed
that the target almost moves smoothly but abruptly changes over several frames.
However, most trackers \cite{WLMC-PAMI, SAMC-TIP} simply consider all moves as abrupt changes
and thus tend to suffer from drifting in case of background clutter or distractions.

\begin{figure}
  \centering
  \includegraphics[width=\textwidth]{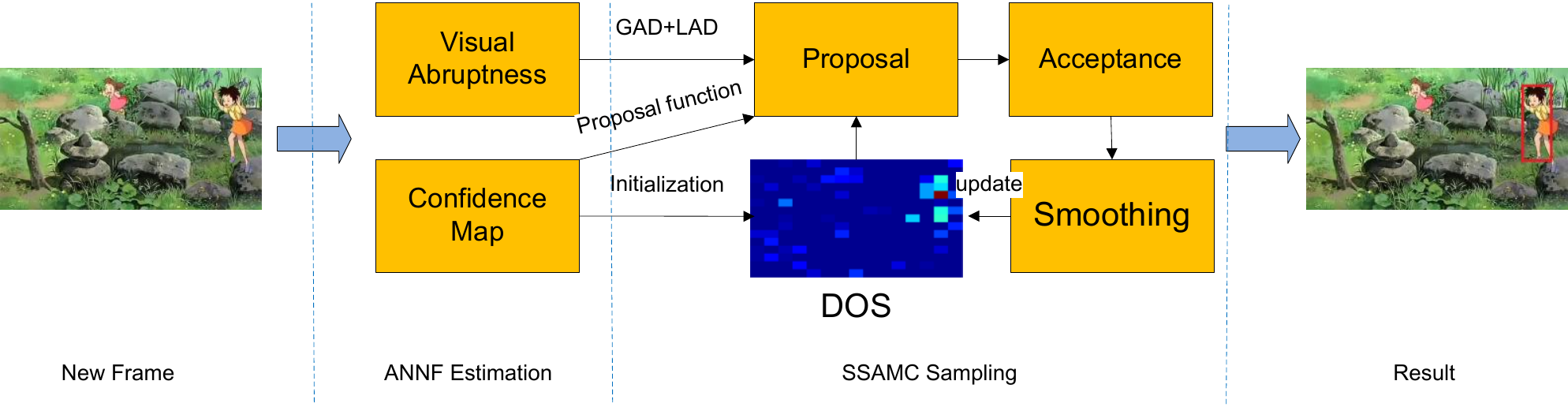}
  \caption{Flow chart of the proposed tracker. The confidence and the visual abruptness
  guide the sampling around the posterior modes, and the sampler iteratively updates the density-of-states to achieve good estimation.}
  \label{fig:system}
\end{figure}

To overcome these difficulties,
we present a novel stochastic sampling method for abrupt motion tracking.
First, we utilize an approximate nearest neighbor field (ANNF) algorithm to compute the importance proposal probabilities,
which drive the Markov chain dynamics and achieve tremendous speedup in comparison with previous methods \cite{WLMC-PAMI, SAMC-TIP}.
Second, we incorporate the ANNF into a smoothing stochastic approximation Monte Carlo (SSAMC) framework.
Within the framework, we consider that adjacent subregions probably bear similar likelihood to the target template,
 and accordingly develop a smoothing weight-updating step to distribute the information in each candidate to its neighborhood.
The smoothing step not only improves the efficiency of the existing Monte Carlo algorithms,
but makes our tracker robust to the noises in the nearest neighbor field.
Furthermore, to track the abrupt and the smooth motions simultaneously,
we develop an effective abrupt-motion detection scheme to discover the presence of sudden changes during tracking
so that we can adjust the sample space for more efficient sampling.
\figref{fig:system} illustrates an overview of our system, and \algref{alg:flow} describes our algorithm.

Note that the conference version of this work is presented in \cite{zhou2014nearest},
and this article deepens and expands our previous work.
In particular,
1) we present a substantial additional number of discussions and analysis about the previous literature on abrupt motion tracking;
2) we develop an abrupt-motion detection scheme to handle the challenging problem of tracking the abrupt
and the smooth motions simultaneously;
3) we formalize the proposed stochastic sampling algorithm
and bridge the gap between ANNF and the sampler using a weighted trial distribution;
4) we perform various additional experiments to evaluate the effectiveness of our algorithm for tracking.

The remainder of this paper is organized as follows:
we review the related work in \secref{section:related-work}.
In \secref{section:annf}, we generalize the ANNF estimation into abrupt motion tracking,
which is followed by the proposed sampling-based tracker in \secref{section:ssamc}.
The results of experiments and performance evaluation are shown in \sectionref{section:results}.
Finally, we summarize our work with remarks on potential extensions in \sectionref{section:conclusion}.

%% file: related_work.tex
\section{Related Work}\label{section:related-work}

There is a rich literature on visual tracking,
and a full review of it is beyond the scope of this work (some are provided in \cite{Survey-1, yang2011recent}).
Here, we only discuss the relevant work that motivated our paper.

Particle filter(PF) based methods
\cite{perez2002color,nummiaro2003adaptive,LSST,CoupleLayer,zhong2014robust,Li2007cascade,jia2012visual, dou2014robust}
have been proven powerful in dealing with the non-Gaussian
and the multi-modal state space for visual tracking.
Many assume that the object in question moves smoothly between consecutive frames.
Such a simplified assumption may work well in simple lab environment;
however, it would have troubles in tracking the abrupt motions without significant drift in complex scenes.
To address the limitations of PF, Michael \etal \cite{isard1998condensation} incorporate the condensation algorithm into the importance sampling
to track the target in high-dimensional sample space.
Similarly, Vasanth \etal \cite{philomin2000quasi} combine PF with the quasi-random sampling to handle the abrupt changes.
However, both methods are subject to the local-trap problem in abrupt motion tracking.
Su \etal \cite{su2014abrupt} incorporate a visual saliency model into the particle filtering framework.
However, tracking failure will be caused by the background clutter because the saliency cannot be reliably estimated.

Traditional approaches for abrupt motion tracking are based on multi-scale representation \cite{hua2004multi},
 layered sampling \cite{sullivan1999object} and multi-observation model \cite{Li2007cascade}.
Hua \etal \cite{hua2004multi} propose a multi-scale collaborative searching strategy based on the dynamic Markov network.
Sullivan \etal \cite{sullivan1999object} propose to combine observation likelihoods in different scales for accurate Bayesian estimation.
Multi-scale methods can largely reduce the effect of the fast motion and the search space.
However, the down-sampling operation may induce information loss to a certain extent.
Therefore, in \cite{Li2007cascade}, multi-observation model is constructed on the same image space to alleviate the information loss.
While this method shows promising results in face tracking, the off-line learning procedure makes it practically infeasible.


Our work is also related with approximate nearest neighbor field estimation \cite{barnes2009patchmatch, CSH, he2012computing}.
This technique has found recent success in many computer vision areas,
such as large displacement optical flow estimation \cite{LDOF-NNF, bao2014edge},
 and orderless tracking \cite{hong2013orderless}, etc.
The ANNF estimation does not rely on the motion continuity;
hence, it can provide relatively accurate motion information even though there are great changes.

More recently, the idea of using Markov Chain Monte Carlo in the sequential importance re-sampling particle filter
has been widely explored \cite{khan2004mcmc, khan2005mcmc, WLMC-PAMI, SAMC-TIP, wang2012hamiltonian}.
These approaches typically replace the importance sampling in particle filter with a MCMC sampling step,
which is more efficient in high-dimensional spaces.
Our work is partly motivated by the work \cite{WLMC-PAMI, SAMC-TIP}.
Kwon \etal \cite{WLMC-PAMI} propose to utilize the Wang-Landau Monte Carlo(WLMC) sampling method
to deal with the local-trap problem in abrupt motion tracking.
Along with this thread, in Bayesian context,
Zhou \etal \cite{SAMC-TIP} propose an intensively adaptive MCMC sampling method for abrupt motion tracking.
Compared with \cite{WLMC-PAMI}, the posterior distribution can be more effectively estimated by a stochastic approximation process.
However, this method has to explore the whole sample space uniformly with an inefficient preliminary sampling phase.
Moreover, both two methods consider each move as an abrupt motion, which will cause the trackers' failure in cluttered scenes.

We make two main complementary improvements to \cite{WLMC-PAMI, SAMC-TIP}.
First, we propose a novel stochastic sampling method to search for the global optimum state in the large solution space.
The nearest neighbor field is computed between consecutive frames to guide the Markov chain search
and enhance the efficiency in stochastic sampling stage.
%
Second, we leverage an abrupt-motion detection scheme to discover unexpected motions in a sequence
so that our sampler can adjust the search space adaptively.
This enables us to address the tracking problem including both abrupt and smooth motions.


%% file: annf.tex
\section{Generalizing ANNF into Abrupt Motion Tracking}\label{section:annf}

\begin{figure}
  \centering
  \includegraphics[width=0.8\textwidth]{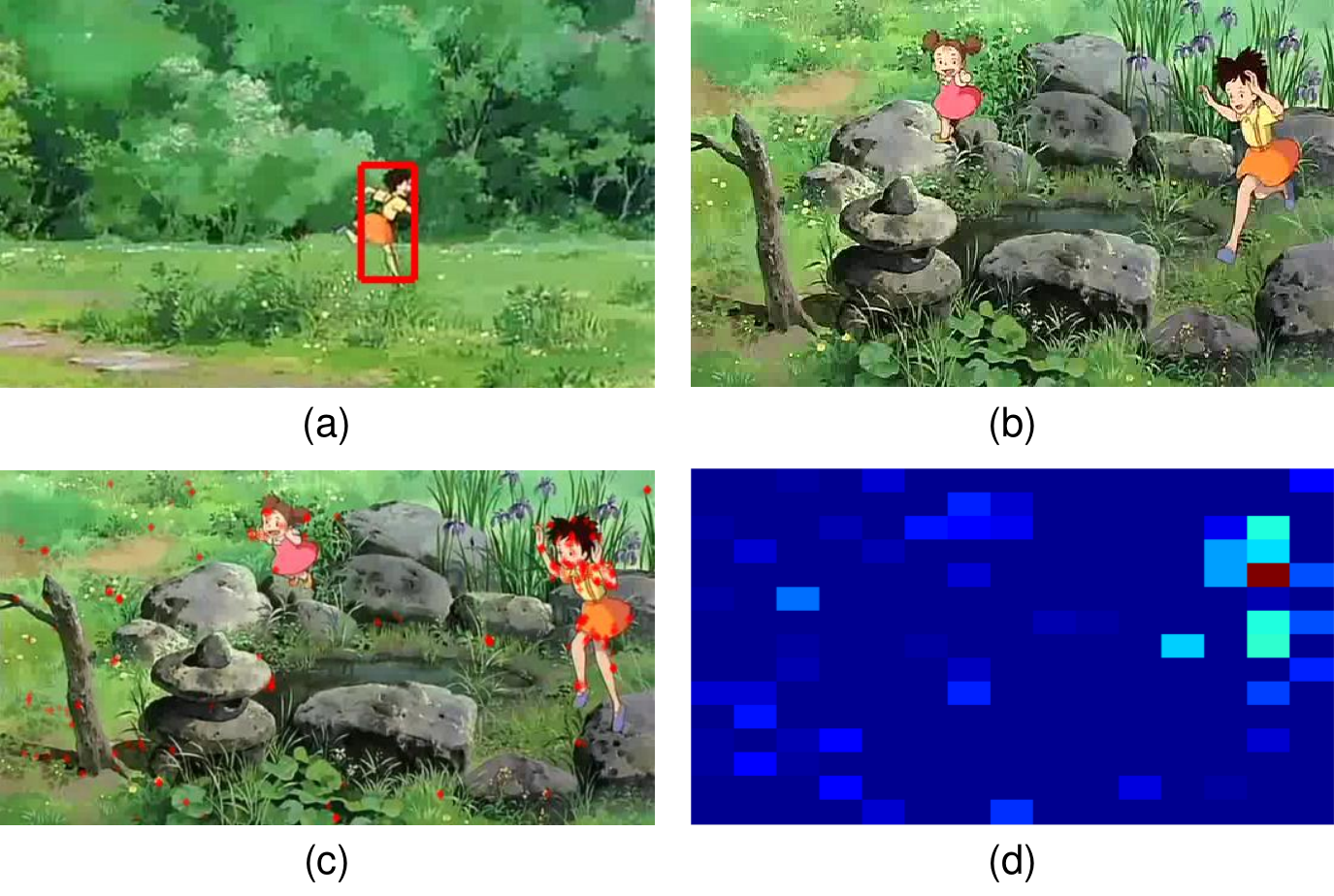}
  \caption{(a)(b)Abrupt motion with shot change; (c)Forward-backward matching result. Each red pixel indicates the center of a patch.
  (d) The confidence map which gives the rough regions that the target might be in.}
\label{fig:annf}
\end{figure}

In this section, we discuss how to generalize the approximate nearest neighbor field estimation into abrupt motion tracking.
A nearest neighbor field between two images is defined as:
for each patch in an image, the most similar patch in another image.
In this work, given two temporally adjacent frames at \(t-1\) and \(t\) (assuming frame \(t-1\) has been tracked),
we discover the rough mode of the target in frame \(t\) using the patch matching method \cite{CSH}.
Additionally, the field enables us to determine whether abrupt motions occur in frame \(t\).

\noindent
\textbf{Confidence Map}\hspace{3mm}
Inspired by the occlusion reasoning scheme in \cite{Bro10e},
 we employ a forward-backward consistency check of the correspondence to achieve more accurate field.
In particular, let \(p_{t-1}\) denote the center of a patch within the bounding box in frame \(t-1\),
the forward matching patch in frame \(t\) is denoted as its center \(q_t\);
the backward matching patch corresponding to \(q_t\) is \(s_{t-1}\).
In this work, we assume that \(q_t\) is reliably estimated
if the backward matching patch \(s_{t-1}\) belongs to the bounding box in frame \(t-1\).
By removing the unreliable correspondences,
 we obtain a set of patches \(\mathcal{O}_t\)
which are viewed as the promising regions where the target might be in the \(t\)-th frame.
Then, the confidence of a pixel \(o\) in \(\mathcal{O}_t\) is estimated according to its \(\textit{incoherence}\)\cite{CSH},
which is defined as the pixel numbers \(M_{t-1}^{o}\) at time \(t-1\) that \(o\) is mapped to,
as computed by,
\begin{equation}\label{eq:incoherence}
  H_{t}^{o} =
  \begin{cases}
    M_{t-1}^{o} & \text{if} \quad o\in\mathcal{O}_t, \\
    0 & \text{otherwise}
  \end{cases}
\end{equation}
where \(H_t^{o}\) is the incoherence of the pixel \(o\) at time \(t\).
Given the incoherence map, the confidence map (\figref{fig:annf}(d)) is obtained by a quantization process
 on \(m\) disjoint subregions (\secref{subsection:via-ssamc}),
\begin{equation}\label{eq:confidence}
  \lambda_t^i = \frac{\sum_{o=1}^{N_t^i}H_{t}^{o}}{N_t^i}
\end{equation}
where \(\lambda_t^i\) is the confidence of the \(i\)-th subregion at time \(t\) and \(N_t^i\) indicates the pixel number in it.

\noindent
\textbf{Abrupt-motion Detection}\hspace{3mm}
To make our algorithm robust in both abrupt and smooth scenarios,
we propose two criteria which can well evaluate the abrupt degrees of the target and the background.

\(\bullet\) \textit{Global Abrupt Degree(GAD)}.
This criterion evaluates how much a frame has changed in comparison with previous frame.
It is computed with the consideration that the matching error image represents the variance of the pixels between consecutive frames.
However, we observe that the matching errors in high-frequency regions, \ie the edges, are always large when using \cite{CSH}
which misleads the abrupt-degree estimation.
Thus, we refine the error image \(\Gamma_t\) at time \( t\) using the edge map \(D_t\) to get a refined error image \(R_t\),
\begin{equation}
  R_t = \frac{\Gamma_t}{\text{DIL}(D_t)+A}
\end{equation}
where \(A\) is an all-one matrix and \(\text{DIL}(\cdot)\) denotes a dilation operator with a specific structuring element object,
 that is, a \(3\times3\) all-one matrix.
Then, the global degree \(g_t\) is computed by,
\begin{equation}
  g_t = \frac{\sum_{i=1}^{U_t}R_t^i}{\max(R_t)\times U_t}, \quad \max(R_t) = \max(R_t^1, R_t^2, \ldots, R_t^{U_t})
\end{equation}
where \( U_t \) denotes the number of pixels in frame \(t\),
and \(R_t^i\) indicates the error value of the \(i\)-th pixel.

\(\bullet\) \textit{Local Abrupt Degree(LAD)}.
While the global abrupt degree is effective under camera switching conditions,
 it leads to unsatisfactory estimation results in other cases, \ie fast motion.
The limitation is overcome by the local measurement.
Given the pixels \(\mathbi{p}\) in the bounding box at time \(t-1\) and the matching set \(\mathbi{q}\) at \(t\),
we model them with \(K\)-component Gaussian mixture model(GMM), respectively,
\begin{equation}
  \text{Pr}_{t-1}^\mathbi{p}(x) = \sum_{k=1}^{K}\pi_{k}\mathcal{N}(x|\mu_{k}, \Sigma_{k}), \quad
  \text{Pr}_{t}^\mathbi{q}(x) = \sum_{k=1}^{K}\pi_{k}'\mathcal{N}(x|\mu_{k}', \Sigma_{k}')
\end{equation}
where the \(k\)-th component in the sample distribution \(\text{Pr}_{t-1}^\mathbi{p}(x)\) is characterized by
normal distribution \(\mathcal{N}(x|\mu_k, \Sigma_k)\) with weight \(\pi_k\), mean \(\mu_k\)
and covariance matrix \(\Sigma_{k}\), and it is similar in \(\text{Pr}_{t}^\mathbi{q}(x)\).
Note that the representation is supported by the theory
 that each single pixel can be presented in a distribution as a \(\delta\) function
 which can be generally written as a Gaussian distribution with zero covariance.
We now use the Hellinger distance\cite{kristan2011multivariate}
to compute the similarity between the two GMMs.
Here, the distance is to measure the local abrupt degree \(l_t\) in the \(t\)-th frame,
\begin{equation}
  l_t^2  = 1 - \int\sqrt{\text{Pr}_{t-1}^\mathbi{p}(x)\times\text{Pr}_{t}^\mathbi{q}(x)}dx
\end{equation}

Given the global degree \(g_t\) and the local degree \(l_t\),
the abruptness \(V_t\) of the target can be computed by,
\begin{equation}\label{eq:v}
  V_t =
  \begin{cases}
    1 & \text{if} \quad  g_t > T, \\
    \text{sign}(a_t-0.5)& \text{otherwise}
  \end{cases}
\end{equation}
where \(T\) is empirically set to \([0.1, 0.2]\) and \(a_t = 0.55 + g_t\times(l_t-0.45)\).
Here, frame \(t\) is considered as an abrupt frame if \(V_t\) equals to one.

Note that although the ANNF provides valuable prior information about the target's movements,
a na{\"i}ve embedding of the field into visual tracking requires the consideration of several factors,
such as the noises in the field, the drifting problem due to short-term correspondence, etc.
In this study, we use a smoothing stochastic sampling Monte Carlo algorithm (\secref{section:ssamc}), which is robust to noise,
to estimate the accurate state of the target from a noisy nearest neighbor field.

%% file: ssamc.tex
\section{Stochastic Sampling Based Tracking Algorithm}\label{section:ssamc}

\subsection{Bayesian Formulation}
In this work, visual tracking is formulated as a dynamic Bayesian inference task
in hidden Markov model.
Let \(X_t = \{x_t,y_t,s_t\}\) represent the state of a target at time \(t\), where \((x_t, y_t)\)
indicates the 2D coordinate of the target in the image plane, and \(s_t\) denotes its scale.
Given the observations \(Z_{1:t}=\{z_1,z_2,\cdots,z_t\}\) up to the \(t\)-th frame,
we estimate the optimal state of the target at time \(t\) by the \textit{maximum a posterior}(MAP) estimator:
\begin{equation}
  \hat{X_{t}} = \argmax_{X_{t}^{i}} p(X_{t}^{i}|Z_{1:t})
\end{equation}
where \(X_t^i\) indicates the \(i\)-th sample at time \(t\).
According to the Bayes theorem, the posterior distribution \(p(X_t|Z_{1:t})\) can be estimated recursively by:
\begin{equation}\label{equation:bayesian}
  p(X_t|Z_{1:t})\propto p(Z_t|X_t)\int p(X_t|X_{t-1})p(X_{t-1}|Z_{1:t-1})dX_{t-1}
\end{equation}
where \(p(X_t|X_{t-1})\) is the motion model
that describes the evolution of the state variable, and
\(p(Z_t|X_t)\) is the observation model
measuring the similarity between the candidate samples and the appearance model.

\input{algo1}

\subsection{Sampling}
\label{section:sampling}
Directly sampling from the filtering distribution \(p(X_{t}|Z_{1:t})\) is intractable
since it is not a simple, standard distribution considered so far.
However, as is often the case, we are able to evaluate the desired distribution
for any given sample up to the normalizing constant \(\mathcal{Z}_p\).
Without loss of generality, in this sub-section,
we use \(p(x)\) to represent the filtering distribution for convenience.
Then, we can write the distribution in the following form:
\begin{equation}
  p(x) = \frac{1}{\mathcal{Z}_{p}}\tilde{p}(x),\; x\in\mathcal{X}
\end{equation}
where the density \(\tilde{p}(x)\) gives the unnormalized probability of a state,
and can be readily evaluated; \(\mathcal{X}\) indicates the sample space.
In physics and chemistry, \(\tilde{p}(x)\) is commonly called the Boltzmann factor:
\begin{equation}
  \tilde{p}(x) = \exp(-E(x)),\; x\in\mathcal{X}
\end{equation}
where \(E(x)\) is an energy function.

\subsection{Metropolis-Hastings Algorithm}
\label{section:mh}
The Metropolis-Hastings algorithm is now widely used to estimate the filtering distribution
because it can draw samples from probability distribution \(p(x)\),
if we can easily evaluate the value of \(\tilde{p}(x)\).
Note that it is not necessary to calculate \(\mathcal{Z}_p\) which is often difficult in practice.
As more samples are proposed, the distribution of these samples more closely approximates the desired distribution \(p(x)\).
The algorithm is performed in two steps:
\begin{itemize}
  \item \textbf{Proposal Step}
    Draw a candidate state \(X_{t}'\) from a proposal function \(Q(X_{t}'|X_{t})\) given the current state \(X_{t}\).
    The function \(Q\) is commonly designed based on a motion transition model, \ie the Gaussian distribution.
  \item \textbf{Acceptance Step}
    Compute the acceptance probability \(\alpha\),
    which is used to determine whether to accept or reject the candidate:
    \begin{equation}
      \alpha(X_{t}';X_{t}) = \min\{1, \frac{p(X_{t}'|Z_{1:t})Q(X_{t};X_{t}'))}{p(X_{t}|Z_{1:t})Q(X_{t}';X_{t}))}\}
      \label{equation:mh}
    \end{equation}

\end{itemize}

While the Metropolis-Hastings based tracking algorithms\cite{khan2005mcmc, benfold2011stable}
work well in some cases,
it is prone to get trapped in the local energy maxima
when the energy landscape of the state space is rugged.
In the next section, we introduce a novel stochastic sampling method to alleviate this problem.

\subsection{Abrupt Motion Tracking via Stochastic Sampling}\label{subsection:via-ssamc}
As discussed in \sectionref{section:sampling},
the Boltzmann distribution can help predict the probability distribution for the energy function \(E\).
However, once the abrupt motions occur, the energy landscape will be rugged, and this will cause the local-trap problem.
To address this issue, we partition the image space \(\mathcal{X}_t\) at time \( t\) into \( m \) disjoint subregions
\( \mathcal{X}_t = \bigcup_{i=1}^m E_i \) according to the energy function \( E(x) \).
 Here, \( E(x) = p(X_t | Z_{1:t})\) is the posterior probability in \equationref{equation:bayesian}.
Then, we design an effective sampler to simulate a random walk in the subregions
 so that the motion uncertainty can be captured.

Instead of the original posterior distribution,
we construct a novel trial density function, called weighted trial function, for importance sampling,
\begin{equation}\label{equation:trial}
  p_{\omega}(X_t) \propto \sum_{i=1}^m\lambda_t^i\frac{p(X_t|Z_{1:t})}{\omega_i}I(X_t\in E_{i})
\end{equation}
where \(\lambda_t^i\) is the confidence of the \(i\)-th subregion at time \(t\),
 which controls the sampling frequency of this subregion;
 \(I(\cdot)\) is the indicator function, and \(\omega_i = \int_{E_i}p(X_t|Z_{1:t})dX_t\)
 is called density-of-states (DOS) of the distribution.
It has been demonstrated that if we are able to estimate the density term for each subregion,
sampling from \(p_{\omega}(X_t)\) will lead to a random walk in the image space (by regarding each subregion as a point) \cite{liang2007stochastic}.
Hence, the local-trap problem can be overcome.
Compared with \cite{SAMC-TIP},
the weight parameter \( \lambda_t^i \) in the trial distribution
 controls the similarity between the target distribution, i.e., \(p(X_{t}|Z_{1:t})\)
and the trial distribution \(p_{\omega}(X_{t})\).
Clearly, one can incorporate any priors into the trial function by adjusting the weight parameter.
Here, the ANNF is utilized to conduct the sampler to coverage fast to the posterior distribution.

In what follows, we elaborate the three major stages in our sampling-based tracking algorithm, which are,
the proposal step, the acceptance step and the smoothing step.

\subsubsection{Proposal Step}\label{subsubsection:proposal}
The choice of the proposal function is significant to our algorithm.
For continuous sample space, a common choice is the Gaussian distribution centered on the current state,
leading to an important trade-off in determining the variance of this proposal function.
\cite{WLMC-PAMI, SAMC-TIP} use a large variance to capture large motions,
in which case, a large percentage of samples are drawn from the unpromising regions, thereby decreasing the acceptance rate.
In this work, we observe that the states with high posterior probability should be more frequently sampled, and vice versa.
Therefore, we develop an adaptive proposal function based upon the ANNF estimation (\sectionref{section:annf}).
The field provides importance probabilities for the image subregions
 to make the samples in promising regions be proposed with higher probabilities.

Our proposal function includes two basic moves: \textit{global random walk} and \textit{local random walk}.
Notably, 1) we perform global random walk on the image subregions \(\mathcal{X}_t\) to account for the large motion uncertainty.
At each move, a subregion \( E_i \) will be selected with probability \(\rho\),
\begin{equation}
  {\rho}(E_{i}) =
  \begin{cases}
    \theta & \text{if} \quad \lambda_t^i > 0, \\
    1-\theta & \text{otherwise}
  \end{cases}
  \label{equation:cp}
\end{equation}
where \( \rho(E_i) = \theta\) if \(E_i\) contains at least one patch in the ANNF of frame \(t\).
After selecting a subregion, a candidate pixel is uniformly determined within it;
2) we also perform a Gaussian random walk to explore the local sample space
whose step size varies according to the normal distribution.
The local random walk tends to propose the states close to the previous one
since the target generally moves smoothly.

With the aforementioned notations, our proposal distribution can be formulated into a mixture model,
\begin{equation}\label{eq:q}
  Q(X_{t}';X_{t}) = \beta\mathcal{N}(X_{t}';X_{t},\Sigma) + (1-\beta)Q_{a}(X_{t}';X_{t})
\end{equation}
where \(\mathcal{N}(\cdot;X_{t}, \Sigma)\) is a normal distribution with mean \(\mu=X_t\) and a small variance \(\Sigma\);
\(Q_a(X_{t}';X_t)\) is an adaptive proposal function,
\begin{equation}
  Q_{a}(X_{t}';X_{t}) \propto \rho(E_{j}), \hspace{2mm} \text{where}\hspace{2mm} X_{t}'\in E_{j}
\end{equation}
The parameter \(\beta\in[0,1]\) in \equref{eq:q} balances the proposal between the global random walk and the local random walk.

To track the abrupt and the smooth motions simultaneously, our proposal function
adaptively adjusts the candidate sample space \(\mathcal{X}_t'\) according to the abruptness of the frame,
\begin{equation}
  \mathcal{X}_t' =
  \begin{cases}
    \mathcal{L}(\hat{X}_{t-1})& \text{if} \quad V_t = 0, \\
    \mathcal{X}_t & \text{otherwise}
  \end{cases}
  \label{eq:va}
\end{equation}
where \(\hat{X}_{t-1}\) denotes the best state in previous frame,
and \(\mathcal{L}(\cdot)\) indicates the nearby regions of a state, \ie \(5\times5\) neighborhood.
Note that this adaptive proposal always biases the sampling towards the promising regions
 to improve the sampling efficiency as well as the accuracy of the state estimation.

\subsubsection{Acceptance Step}
Suppose that a candidate sample \(X_{t}'\) has been generated using the proposal function in \equationref{eq:q},
accepting it or not is determined by the Metropolis-Hastings rule,

\begin{equation}\label{eq:acceptance}
  \begin{split}
    \alpha(X_{t}';X_{t}) & = \min\left\{1, \frac{p_{\omega}(X_{t}')}{p_{\omega}(X_{t})} \frac{Q(X_{t};X_{t}')}{Q(X';X_{t})}\right\} \\
           & = \min\left\{1, \frac{P(X_{t}'|Z_{1:t})\frac{\lambda_t^{J_{X_t'}}}{\omega_{J_{X_{t}'}}}Q(X_{t};X_{t}')}{P(X_{t}|Z_{1:t})
              \frac{\lambda_t^{J_{X_t}}}{\omega_{J_{X_{t}}}}Q(X_{t}';X_{t})}\right\}
  \end{split}
\end{equation}
where \(J_{X_t}\) denotes the index of the subregion containing \(X_t\).
Different from \cite{WLMC-PAMI, SAMC-TIP}, the density of each subregion is initialized with its confidence,

\begin{equation}\label{eq:init}
  \omega_i = \exp(-\tau\times\lambda_t^i)
\end{equation}
where \(\tau\) is empirically set to \(1000\) in our experiments.

Our acceptance ratio in \eqref{eq:acceptance} has two advantages compared to that in \cite{WLMC-PAMI, SAMC-TIP}.
The first is that the acceptance ratio and the density initialization procedure enable us to escape from the local maxima and reach the global maximum.
At a local maximum, the ratio \(\frac{\lambda_t^i}{\omega_i}\) initially has a larger value than that at the global maximum
because the confidence value \(\lambda_t^i\) is smaller, and the DOS term \(\omega_i\) is larger according to \eqref{eq:init}.
Hence, the samples at the local maximum are more easily rejected compared with those at the global maximum.
While the simulation goes on, the ratio will further decrease because the DOS will increase(\secref{section:smoothing}).
By contrast, the ratio at the global maximum will increase.
This process helps our algorithm escaping the local maxima;
second, during tracking, the confidence value \(\lambda_t^i\) in \eqref{eq:acceptance}
always drives our sampler to accept the candidate samples in the promising regions.
This largely reduces the rejection rate and enhances the sampling efficiency.

\subsubsection{Smoothing Step}\label{section:smoothing}
The success of stochastic approximation Monte Carlo(SAMC) algorithm \cite{liang2007stochastic} depends crucially on the self-adjusting mechanism,
which enables the sampler to explore the entire image space.
However, the density learning method in SAMC has not yet reached the maximal efficiency
since it ignores the difference between the neighboring and the non-neighboring regions.
Intuitively, a sample \(X_t\) may contain some information of the neighboring regions.
For instance, if \(X_t\) in subregion \(E_i\) is rejected,
the samples in the neighborhood will be probably rejected as well,
and vice versa.
Accordingly, we improve the density learning method by including a smoothing step at each iteration.

More specifically, in the proposal step,
we allow multiple samples to be generated at each iteration and employ a smoothed estimator
\(f_i^k\) when updating the density-of-state term, where \(f_i^k\) is the probability that a sample can be drawn from the subregion \(E_i\) at iteration \(k\).
Let \(X_{t}^{(k,1)}, X_{t}^{(k,2)}, \ldots, X_{t}^{(k,n)}\) be \(n\) samples generated in the proposal step at iteration \(k\) in frame \( t\).
Since \(n\) is usually a small number(\(n=5\) in our experiment),
the samples form a sparse frequency vector \(\textbf{r}^k =(r_1^k, r_2^k, \ldots, r_m^k) \) with \(r_i^k = \sum_{j=1}^nI(X_t^{(k,j)}\in E_i)\).
It is worth mentioning that \(r_i^k/n\) is not a good estimation for \(f_i^k\) because the law of large numbers does not serve in this situation.
Since the image space is partitioned smoothly in this paper,
we assume that information in nearby regions can help produce more accurate estimate of \(f_i^k\).
Therefore, we improve the frequency estimator with a smoothing method, that is, the Nadaraya-Waston kernel estimator \cite{liang2009improving},
\begin{equation}\label{eq:smooth}
  f_i^k = \frac{\sum_{j=1}^{m}W(M(i-j))r_j^k/n}{\sum_{j=1}^{m}W(M(i-j))}
\end{equation}
where \(M(i-j)\) measures the Euclidean distance between the centers of subregions \(E_i\) and \(E_j\).
\(W(\cdot)\) is a double-truncated Gaussian kernel function to control the smoothing scope,
\begin{equation}
  W(z) =
  \begin{cases}
    \exp(-z^{2}/2) & \text{if} \quad |z| < C, \\
    0 & \text{otherwise}
  \end{cases}
  \label{equation:w}
\end{equation}
where \(C\) is empirically set to \(100\).
After achieving the smoothed estimation \(\textbf{f}^k = (f_1^k, f_2^k, \ldots, f_m^k)\),
we update the density-of-state of \(E_i\) as,
\begin{equation}\label{eq:self-adjust}
  \omega_i^{k+1} = \omega_i^k + \exp(\gamma_k(f_i^k-\pi_i)), \quad i = 1, 2, \ldots, m
\end{equation}
where \(\omega_i^k\) indicates the DOS term of \(E_i\) at iteration \(k\);
\(\pi=(\pi_1,\pi_2,\ldots,\pi_m)\) is a vector with \(0<\pi_i<1\) and \(\sum_{i=1}^m\pi_i=1\),
which defines the desired sampling frequency of each subregion;
\(\gamma_k=\frac{k_0}{\max(k_0,k)}\)(\(k_0\) is a pre-specified constant) is a gain factor controlling the updating speed of the density-of-states.

The smoothing weight-updating step has more superiorities in comparison to the existing algorithms:
1) suppose a candidate is rejected in the acceptance step,
the density-of-states of the subregions that the candidate belongs to and nears with will be adjusted to a larger value.
Thus, in the next iteration, our algorithm can jump out from these subregions with a high probability.
This is important for our approach not to fall into the local maxima;
2) by distributing the information contained in a subregion to the nearby ones,
the smoothing scheme in the weight-updating step not only improves the accuracy of the DOS estimation,
but makes our method robust to the noises in the ANNF.


%% file: algo1.tex
\begin{algorithm}
  \begin{algorithmic}[1]
    \STATE \textbf{Input:} Image frames \( F_1, F_2, \cdots, F_T \); object state \(\hat{X_1}\) in the first frame; iteration number \(K\); sample
number \(N\) in each iteration.
    \STATE \textbf{Output:} Tracking results \( \{X_t\}_{t=2}^T \).
    \STATE \textbf{Initialize:}
      \STATE\hspace{\algorithmicindent} Foreground and background HSV histograms generation: \(h_F\) and \(h_B\)
      \STATE\hspace{\algorithmicindent} Image space partition: \( \mathcal{X} = \bigcup_{i=1}^m E_i \)
    \STATE \textbf{Main Procedure:}
    \FOR{\(t\) = 1 to \(T\)}
      \STATE \textit{//ANNF Stage}
      \STATE Estimate the confidence map \(\lambda_t\) using \equref{eq:confidence};
      \STATE Detect the presence of abrupt motions according to \equref{eq:v}.
      \STATE \textit{//SSAMC Stage}
        \STATE Determine the sample space \(\mathcal{X}_t'\) according to \equref{eq:va};
        \FOR{\(k\) = 1 to \(K\)}
          \FOR{\(n\) = 1 to \(N\)}
            \STATE (\textit{Proposal}) Propose a candidate sample \(X_t'\) using \equref{eq:q};
            \STATE (\textit{Acceptance}) Accept \(X_t'\) with probability \(\alpha\) in \equref{eq:acceptance};
          \ENDFOR
          \STATE (\textit{Smoothing}) Calculate \(\textbf{f}^k = (f_1^k, f_2^k, \ldots, f_m^k)\) in \equref{eq:smooth};
          \STATE Update the density-of-states according to \equref{eq:self-adjust};
        \ENDFOR
      \STATE \textit{//Inference}
      \STATE MAP estimate to get \(\hat{X}_t\).
    \ENDFOR
  \end{algorithmic}
  \caption{The proposed abrupt motion tracking algorithm.}
 \label{alg:flow}
\end{algorithm}

%% file: experiments.tex
\section{Experimental Results}\label{section:results}

\subsection{Experiment Setup}
The proposed tracker is implemented in MATLAB and runs at 2fps on a PC\@.
Given an image sequence, we manually label the state of the target in the first frame.
For brevity, we will refer our method as SSAMC from now on.

In this work, we utilize the color-based appearance model \cite{perez2002color}.
The foreground and the background are represented with two HSV histograms \(h_F\) and \(h_B\), respectively.
The number of bins in each channel is equally set to 10.
For each candidate \(X_t^k\),
we estimate the similarity between the state and the templates with Bhattacharyya metric,
\(d_F = D(X_t^k, h_F)\) and \(d_B = D(X_t^k, h_B)\).
Finally, the likelihood function is formulated as
\(  p(Z_{1:t}|X_{t}^{k}) = \frac{1}{1+\exp(d_{F}-d_{B})} \).

In our experiments, we set the patch size to \(8\times 8\) for the ANNF estimation.
The image space is empirically partitioned into \(15 \times 15\) disjoint subregions
according to the energy function \(E_x\).
The proposal variance in \equationref{eq:q} is set to \(\Sigma = diag(\sigma_{x}^{2}, \sigma_{y}^{2}, \sigma_{s}^{2})
= diag(8.0^{2}, 4.0^{2}, 0.013^{2})\) in which \(\sigma_x\) and \(\sigma_y\) denote the variances of 2D coordinate,
and \(\sigma_s\) denotes the variance of target scale.
In \equationref{equation:cp}, \(\theta\) is empirically set to \(0.8\).
\(\beta\) in \equationref{eq:q} is set to \(0.2\) to facilitate the global exploration of the proposal function.
\(k_0\) in the gain factor is set to \(N/4\) where \(N\) is the number of samples.
In every experiment, the iteration number \(K\) for sampling is \(120\) and in each iteration, we propose \(n=5\) samples.
The desired sampling distribution is set to be uniform, \ie \(\pi_i = 1/m, i = 1,2,\ldots,m\).
For fair evaluation, in \equationref{eq:v}, we fix the parameter \(T=0.2\) in our experiments,
 although slightly different values for different videos can produce better results.

\input{dataset}

To evaluate our algorithm,
 we selected 6 typical image sequences with various abrupt motion properties from \cite{WLMC-PAMI}.
Details about the sequences are listed in \tableref{table:dataset}.
We compare the proposed method with other 6 state-of-the-art algorithms:
WLMC \cite{WLMC-PAMI}, SAMC \cite{SAMC-TIP}, SCM \cite{zhong2014robust},
VTD \cite{kwon2010visual}, LSST \cite{LSST}, and saliency-based particle filter (referred as SaPF) \cite{su2014abrupt}.
For fair comparison, we run the source codes provided by the authors with tuned parameters to obtain their best performance.

\subsection{Quantitative Evaluation}
\input{cle}
\input{vor}
\input{cle-smooth}

\begin{figure}[t]
  \centering
  \includegraphics[width=\textwidth]{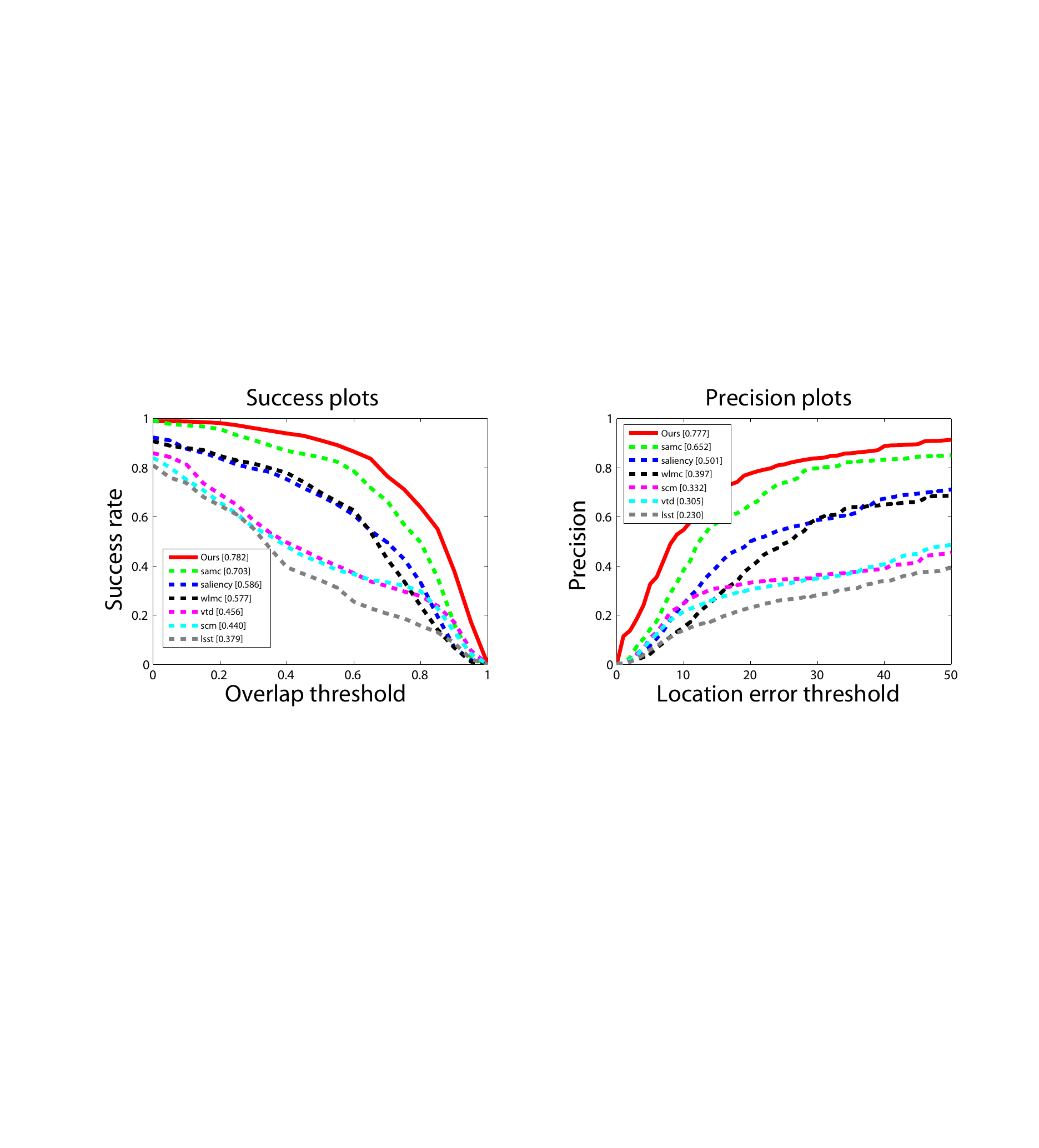}
  \caption{Precision Plots and Success Plots. The overall performance score for each tracker is shown in the legend.}
  \label{fig:plots}
\end{figure}
\begin{figure}[t]
  \centering
  \includegraphics[width=\textwidth]{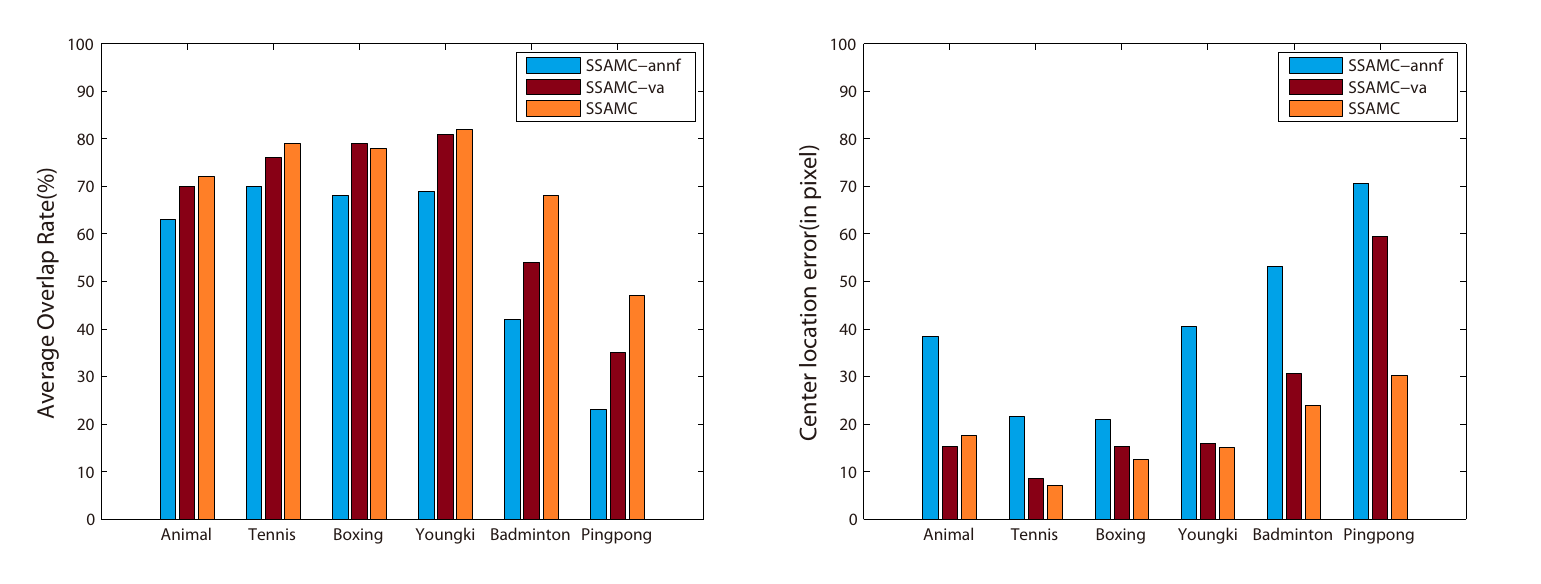}
  \caption{Overlap rates(\%) and location errors(in pixel) of SSAMC-annf, SSAMC-va and SSAMC on the sequences.}
  \label{fig:bar}
\end{figure}

1) \textit{Performance of the Tracking Algorithms:}
Two widely used criteria have been employed in this paper to evaluate the performance of the trackers:
1) \textit{Center Location Error(CLE)} that evaluates the position errors
 between the centers of the tracking results and those of the ground truth;
2) \textit{VOC Overlap Ratio(VOR)} that measures the success ratio of the algorithms,
which is calculated by \(VOR = \frac{|B_r\cap B_g|}{|B_r\cup B_g|}\),
where \(B_r\) denotes the tracked bounding box, \(B_g\) is the ground truth box and
\(\left|\cdot\right|\) denotes the number of pixels in a region.
Besides, the average CLE and the average VOR are calculated on each sequence to evaluate the overall performance of our tracker.

\tableref{table:cle} and \tableref{table:vor} respectively summarize the average CLE and the average VOR of all the six trackers on all 6 sequences.
The potential benefits of our tracker are notable:
 it performs best on 5 of 6 videos in terms of the average CLE,
 and 4 of 6 videos in terms of the average VOR\@.
Besides, it outperforms other trackers by the smallest average CLE and the largest average VOR over all the image sequences.
The performance improvement is particularly impressive in the \textit{Animal}, \textit{Tennis}, \textit{Badminton} and \textit{Pingpong} sequences.
In the \textit{Animal} and \textit{Tennis} sequence,
 the targets move rapidly with unpredictable directions and distances.
Our tracker benefits greatly from the approximate nearest neighbor field that makes the samples be drawn from the promising regions.
Besides, the sequences \textit{Badminton} and \textit{Pingpong} mainly consist of smooth motions.
In this situation, our tracker is more flexible because of the abrupt-motin detection scheme,
while in \cite{WLMC-PAMI} and \cite{SAMC-TIP}, the large sampling variance causes severely tracking accuracy decrease of the smooth motions.
In sum, our tracker outperforms other trackers on most sequences,
 although it shows slightly poor performance on \textit{Boxing}(in terms of VOR) and \textit{Youngki}.
We attribute this to the fact that we have not considered scale changes of the target in this article;
therefore the overlap rate will be a little inaccurate when the scale of the target frequently changes
in \textit{Boxing} and \textit{Youngki}.

We further employ the Precision Plot and the Success Plot \cite{wu2013online} to evaluate the overall performance of these algorithms,
as illustrated in \figref{fig:plots}.
The precision plot summarizes the percentage of frames whose tracking location is within a given distance \(R\) of the ground truth,
and the success plot presents the ratios of successful frames at the thresholds in \([0,1]\).
Here, the successful frame is defined as the frame whose overlap value is larger than a given threshold \(\tau\)(\ie 90\%).
Obviously, our algorithm performs better than other trackers.
More precisely, in the precision plots, it outperforms SAMC by \(\textbf{12.5\%}\), SaPF by \(\textbf{27.6\%}\) and WLMC by \(\textbf{38.0\%}\),
while in the success plots, it outperforms them by \(\textbf{7.9\%}\), \(\textbf{19.6\%}\) and \(\textbf{20.5\%}\), respectively.
Note that the numerical results are computed with the same scheme as \cite{wu2013online}:
the error threshold \(R\) is set to 20 pixels for ranking in the precision plots;
while in the success plots, the area under curve is utilized to rank the tracking algorithms.

The above-mentioned results show the great ability of our algorithm to track the abrupt and the smooth motions simultaneously.
To further evaluate the performance of our tracker in handling the smooth motions,
 we compare it with other algorithms on \textit{Boxing}, \textit{Youngki}, \textit{Pingpong} and \textit{Badminton} movies.
  To only have smooth motions in sequences \textit{Boxing} and \textit{Youngki},
we reinitialized the states of other tracking approaches to the ground truth
when the abrupt motions occur.
The other two sequences remain unchanged.
As listed in \tableref{table:cle-smooth}, the proposed algorithm outperforms other algorithms
 on these sequences even though we did not reinitialize its states.
This demonstrates the effectiveness of our method in handling smooth motions.

2) \textit{Performance of Abrupt-motion Detection:}
To justify the effectiveness of the abrupt-motion detection in our algorithm,
we construct a new tracker, the SSAMC-va tracker, in which the detection process of abrupt motions is neglected.
Thus, the sample space \(\mathcal{X}'\) in \equationref{equation:cp} is always \(\mathcal{X}\), that is, the entire image space.
The quantitative results are illustrated in \figref{fig:bar}.
SSAMC-va shows worse performance than SSAMC on these sequences, especially on the sequences \textit{Badminton} and \textit{Pingpong}.
The fundamental reason is that the abrupt-motion detection scheme largely improves the accuracy of SSAMC in the smooth motions
while the SSAMC-va tracker easily drifts from the target in the smooth movements due to the background clutter.
Therefore, the abrupt-motion detection method is important to our algorithm,
 especially in the scenarios with plenty of smooth motions and background clutter.

3) \textit{Performance of ANNF Estimation:}
To verify the effectiveness of the approximate nearest neighbor field estimation, we also construct a tracker called SSAMC-annf,
in which no initial motion information is provided to the sampler.
The density-of-states in the sampling stage are initialized with all one
and the proposal probability for each cell is equally set to \(1/m\) where \(m\) is the number of cells.
The quantitative results in \figref{fig:bar} prove the importance of ANNF because SSAMC, with ANNF estimation, shows much better performance
than SSAMC-annf.

\input{cost}

4) \textit{Computational Cost:}
We compare the computational cost of the proposed algorithm with other two sampling-based methods, WLMC \cite{WLMC-PAMI} and SAMC \cite{SAMC-TIP}.
The results, as shown in \tableref{table:cost}, are estimated under \(640\times480\) videos and \(600\) samples.
Note that our runtime is slightly longer than other algorithms mainly because we implement the algorithm in MATLAB language rather than C/C++.
Taking this language factor into account, the proposed tracking algorithm has no additive computational burden compared to the other two methods,
because the ANNF helps largely reduce the search space.
Besides, the smoothing weight-updating scheme improves the convergence rate of the Markov chain.

\begin{figure}[t]
  \centering
  \includegraphics[width=\textwidth]{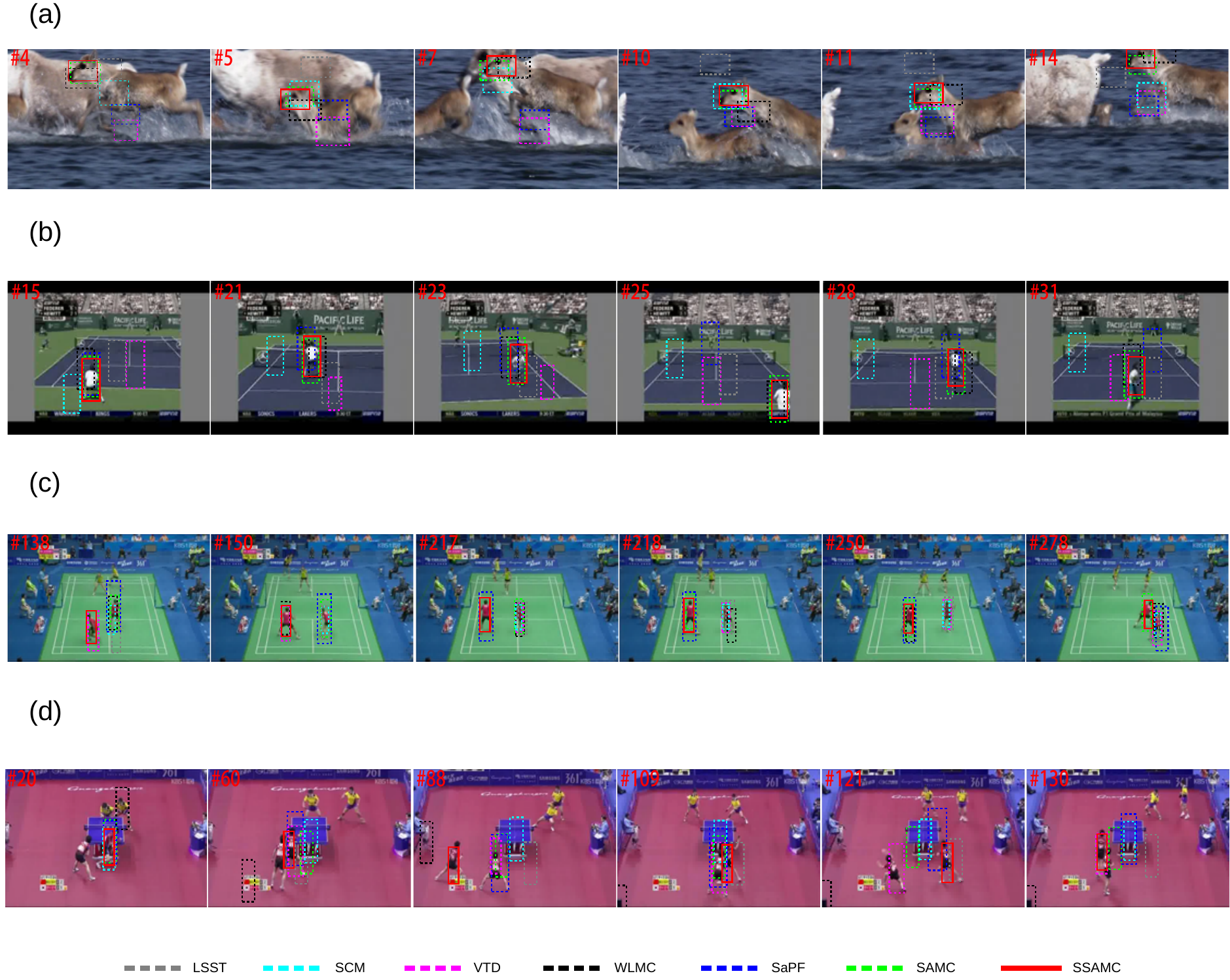}
  \caption{Tracking results on fast motion sequences: (a)\textit{Animal}, (b)\textit{Tennis}, (c)\textit{Badminton}, (d)\textit{Pingpong}.
           Figure best viewed in color.
  }
  \label{fig:results-fastmotion}
\end{figure}
\begin{figure}[t]
  \centering
  \includegraphics[width=\textwidth]{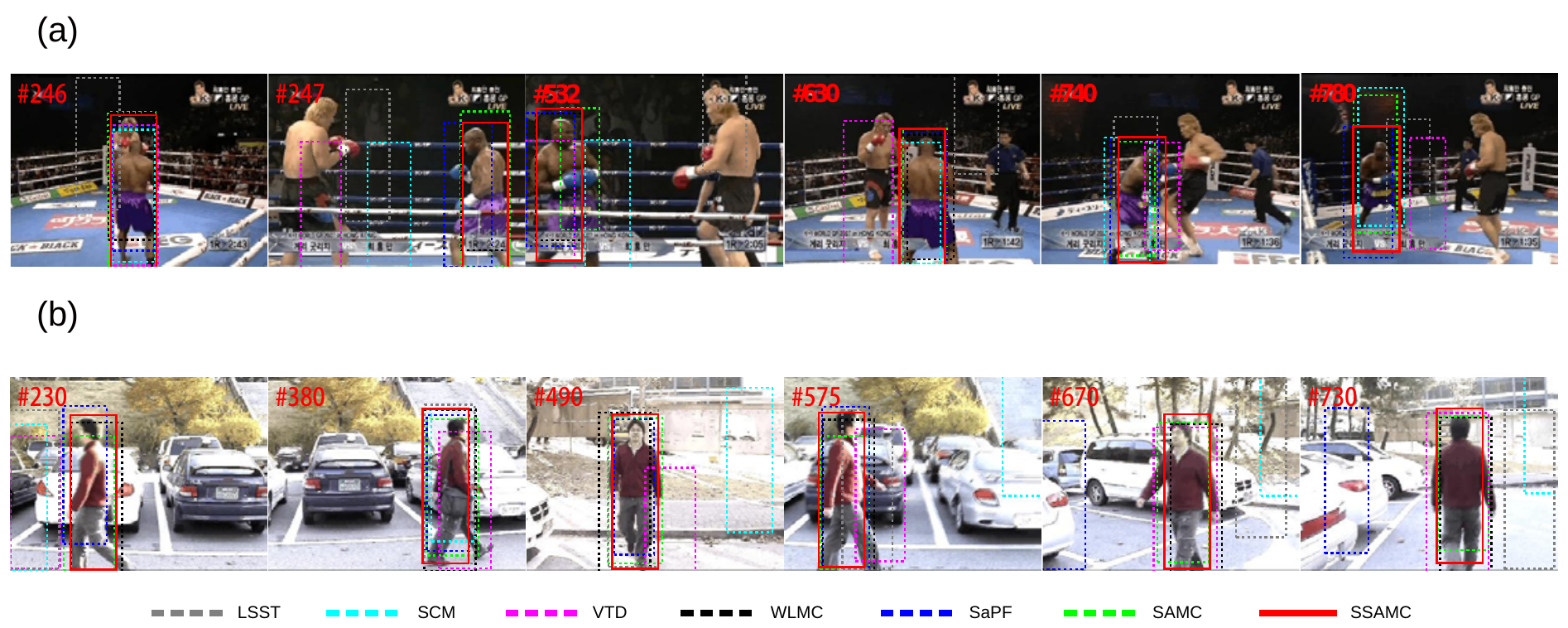}
  \caption{Tracking results on shot change sequences: (a)\textit{Boxing} and (b)\textit{Youngki}. Figure best viewed in color.
  }
  \label{fig:results-shotchange}
\end{figure}

\subsection{Qualitative Evaluation}
In this paper, we categorize the abrupt motions into two classes: \textit{fast motion} and \textit{Shot change}.

1) \textit{Fast Motion:}
We firstly evaluate these trackers on four challenging sequences with fast motion,
which are \textit{Animal}, \textit{Tennis},\textit{Badminton} and \textit{Pingpong}.
The results are illustrated in \figref{fig:results-fastmotion}.
The \textit{Animal} sequence is challenging for tracking as the target moves very fast.
We can see that LSST, SCM, VTD, SaPF and WLMC methods drift from the beginning of the sequence(\ie \#5).
The former three methods lost the target because they heavily depend on the motion continuity;
as for SaPF, it is difficult to estimate accurate saliency map for the head of the animal;
The WLMC tracker requires more samples (more than 1000) to track the target accurately, thus fails in this case.
The SAMC and our method can track the targets successfully through the whole sequence.

In \textit{Tennis} sequence, there are very fast motions and large pose changes of the player.
LSST, SCM, VTD and SaPF trackers fail when the player moves back and forth.
The WLMC tracker is slightly better, but still drifts in several frames (\ie \#23, \#28).
In contrast, SAMC and our trackers are able to track the player throughout the sequence.
As shown in \tableref{table:cle}, our tracker is much better than other trackers in terms of the center location error.

In \textit{Badminton} sequence, the object undergoes heavy occlusion in cluttered background, as well as fast motion in some frames.
Most trackers drift away from the targets because of the interference of similar object in the background.
As shown in \tableref{table:cle} and \tableref{table:vor},
our method is better than others mainly because of the ANNF estimation and the background information included in the appearance model.
The VTD method also performs well with relatively high overlap rates and low center location errors.

In the \textit{Pingpong} sequence, most trackers drift due to the severe occlusion and background clutters(\ie \#88, \#130).
Tracking such an object is extremely challenging because the two red players are difficult to distinguish, even for humans.
As listed in \tableref{table:cle} and \tableref{table:vor}, VTD and our method show significant better performance than other trackers.
WLMC has the lowest overlap rate and the highest location error in this sequence because it drifts at the beginning due to background clutter(\ie \#20).

2) \textit{Shot Change:}
\figref{fig:results-shotchange} shows the tracking results from two challenging sequences to evaluate that whether our method can tackle shot changes
or not.
In the \textit{Boxing} sequence, the target moves smoothly at most time, but the position abruptly changes due to the camera switching.
We can see that LSST, SCM and VTD trackers get lost in tracking the target after the shot changes(\ie \#247).
In contrast, the SaPF, WLMC, SAMC and our approach achieve stable performance on the entire sequence.

In the sequence \textit{Youngki}, a walker undergoes abrupt motions caused by sudden shot changes.
The LSST, SCM and VTD trackers lose the target quickly(\ie \#230) since they can not capture the large motion uncertainty.
SaPF eventually drift in this long-time sequences(\ie \#670).
The WLMC, SAMC and our method track the objects throughout the sequence
because the stochastic sampling scheme help to explore the sample space effectively to capture this type of motions.
The proposed method is slightly worse than SAMC in terms of overlap rates and location errors, however, overall, our results remain acceptable.

%% file: dataset.tex
\begin{table}[!t]
  \centering
  \caption{The sequences and their challenges}
  \label{table:dataset}
  \begin{tabular}{|l|l|l|}
    \hline
    Sequence & Main Challenge & Length \\
    \hline\hline
    Animal & Fast Motion & 15 \\
    \hline
    Tennis & Fast Motion & 31 \\
    \hline
    Boxing & Shot Change & 813 \\
    \hline
    Youngki & Shot Change & 770 \\
    \hline
    Badminton & Fast Motion \& Occlusion & 281 \\
    \hline
    Pingpong & Fast Motion \& Occlusion & 139 \\
    \hline
  \end{tabular}
\end{table}

%% file: cle.tex
\begin{table}[!t]
  \centering
  \caption{Average center location error(in pixel). The best and second best results are shown in red and blue fonts.}
  \label{table:cle}
  \begin{tabular}{|l|c|c|c|c|c|c|c| }
    \hline
    & LSST & SCM & VTD & SaPF & WLMC & SAMC &  SSAMC \\
    \hline \hline
    Animal & 107.59 & 54.17 & 124.30 & 100.69 & 49.70 & {\color{blue}17.94} &  {\color{red}\textbf{17.51}} \\
    Tennis & 75.83 & 105.09 & 102.79 & 45.24 & 36.12 & {\color{blue}19.91} & {\color{red}\textbf{7.01}} \\
    Boxing & 127.72 & 60.62 & 74.92 & 14.52  & 14.08 & {\color{blue}13.00} &  {\color{red}\textbf{12.49}}\\
    Youngki &83.74 & 130.36 & 73.55 & 26.94 & 17.52 & {\color{red}\textbf{12.74}} & {\color{blue}14.99} \\
    Badminton & 53.82 & 50.59 & {\color{blue}32.76} & 34.12 & 46.13 & 44.81 & {\color{red}\textbf{23.99}} \\
    Pingpong & 73.12 & 74.27 & {\color{blue}45.77} & 54.15 & 179.95  & 56.12 & {\color{red}\textbf{30.31}} \\
    \hline \hline
    \textbf{Average} & 86.97 & 79.18 & 75.68 & 45.94 & 57.25 & {\color{blue}27.42} & {\color{red}\textbf{17.72}} \\
    \hline
  \end{tabular}
\end{table}

%% file: vor.tex
\begin{table}[!t]
  \centering
  \caption{Average overlap rate. The best and second best results are shown in red and blue fonts.}
  \label{table:vor}
  \begin{tabular}{|l|c|c|c|c|c|c|c| }
    \hline
    & LSST & SCM & VTD & SaPF & WLMC & SAMC & SSAMC \\
    \hline \hline
    Animal & 0.04 & 0.37 & 0.05 & 0.12 & 0.35 & {\color{blue}0.65} & {\color{red}\textbf{0.72}} \\
    Tennis &0.05 & 0.27 & 0.06 & 0.48 & 0.46 & {\color{blue}0.66} & {\color{red}\textbf{0.79}} \\
    Boxing & 0.12 & 0.38 & 0.34 & 0.75 &{\color{blue}0.78} & {\color{red}\textbf{0.79}} & {\color{blue}0.78} \\
    Youngki & 0.34 & 0.21 & 0.42 & 0.70 & 0.77 & {\color{red}\textbf{0.83}} & {\color{blue}0.82} \\
    Badminton & 0.37 & 0.38 & {\color{blue}0.50} & 0.46 & 0.34 & 0.43 & {\color{red}\textbf{0.68}} \\
    Pingpong & 0.32 & 0.30 & {\color{blue}0.42} & 0.33 & 0.08 & 0.36 & {\color{red}\textbf{0.47}} \\
    \hline \hline
    \textbf{Average} & 0.21 & 0.32  & 0.30 & 0.47 & 0.46 & {\color{blue}0.62} & {\color{red}\textbf{0.71}} \\
    \hline
  \end{tabular}
\end{table}

%% file: cle-smooth.tex
\begin{table}[!t]
  \centering
  \caption{Average center location error(in pixel) when there are only smooth motions. The best and second best results are shown in red and blue fonts.}
  \label{table:cle-smooth}
  \begin{tabular}{|l|c|c|c|c|c|c|c| }
    \hline
    & LSST & SCM & VTD & SaPF & WLMC & SAMC &  SSAMC \\
    \hline \hline
    Boxing & 17.23 & {\color{blue}12.80} & 12.95 & 14.33  & 13.19 & 12.81 &  {\color{red}\textbf{12.49}}\\
    Youngki & 19.01 & 15.02 & {\color{blue}13.22} & 24.40 & 17.35 & {\color{red}\textbf{12.63}} & 14.99 \\
    Badminton & 53.82 & 50.59 & {\color{blue}32.76} & 34.12 & 46.13 & 44.81 & {\color{red}\textbf{23.99}} \\
    Pingpong & 73.12 & 74.27 & {\color{blue}45.77} & 54.15 & 179.95  & 56.12 & {\color{red}\textbf{30.31}} \\
    \hline
  \end{tabular}
\end{table}

%% file: cost.tex
\begin{table}[!t]
  \centering
  \caption{Time Cost of the tracking methods}
  \label{table:cost}
  \begin{tabular}{|l|c|c|c| }
    \hline
    & WLMC & SAMC & SSAMC \\
    \hline \hline
    Cost & 2.4fps & 4fps & 2fps \\
    \hline
    Language  & C/C++ & C/C++ & MATLAB \\
    \hline
  \end{tabular}
\end{table}

%% file: conclusion.tex
\section{Conclusion and Future Work}\label{section:conclusion}

We have proposed a novel stochastic sampling method for abrupt motion tracking in the Bayesian filtering framework.
Within the framework,
the approximate nearest neighbor field estimation is utilized to discover the rough mode of the target at each frame;
after incorporating it into the smoothing stochastic sampling Monte Carlo approach,
our algorithm can more accurately estimate the state of the target.
Moreover, we have developed an abrupt-motion detection scheme so that our tracker can effectively handle both abrupt and smooth motions.
Experiments over several challenging sequences demonstrate the effectiveness of our method compared with other related methods.

In future work, we shall extend the proposed tracking algorithm in three directions:
1) we will firstly improve the method using a robust appearance model (\ie \cite{zhong2014robust, jia2012visual, babenko2011robust});
2) we aim to extend our algorithm to a more efficient one in order to address the abrupt changes in both position and scale;
3) finally, we will expand the method to track the abrupt motions in multi-target scenarios.
Compared with \cite{WLMC-PAMI, SAMC-TIP},
our algorithm can achieve tremendous speedup because the motion fields for all targets can be estimated in a single run.
We will further work on designing an effective data association method to track interactive objects.

%% file: paper.bbl
\begin{thebibliography}{40}
\expandafter\ifx\csname natexlab\endcsname\relax\def\natexlab#1{#1}\fi
\providecommand{\url}[1]{\texttt{#1}}
\providecommand{\href}[2]{#2}
\providecommand{\path}[1]{#1}
\providecommand{\DOIprefix}{doi:}
\providecommand{\ArXivprefix}{arXiv:}
\providecommand{\URLprefix}{URL: }
\providecommand{\Pubmedprefix}{pmid:}
\providecommand{\doi}[1]{\href{http://dx.doi.org/#1}{\path{#1}}}
\providecommand{\Pubmed}[1]{\href{pmid:#1}{\path{#1}}}
\providecommand{\bibinfo}[2]{#2}
\ifx\xfnm\relax \def\xfnm[#1]{\unskip,\space#1}\fi
\bibitem[{Stauffer and Grimson(2000)}]{stauffer2000learning}
\bibinfo{author}{C.~Stauffer}, \bibinfo{author}{W.~E.~L. Grimson},
\newblock \bibinfo{title}{Learning patterns of activity using real-time
  tracking},
\newblock \bibinfo{journal}{{IEEE Transactions on Pattern Recognition and
  Machine Intelligence (TPAMI)}} \bibinfo{volume}{22} (\bibinfo{year}{2000})
  \bibinfo{pages}{747--757}.
\bibitem[{Benfold and Reid(2011)}]{benfold2011stable}
\bibinfo{author}{B.~Benfold}, \bibinfo{author}{I.~Reid},
\newblock \bibinfo{title}{Stable multi-target tracking in real-time
  surveillance video},
\newblock in: \bibinfo{booktitle}{{IEEE Conference on Computer Vision and
  Pattern Recognition (CVPR)}}, \bibinfo{organization}{IEEE},
  \bibinfo{year}{2011}, pp. \bibinfo{pages}{3457--3464}.
\bibitem[{Kim et~al.(2008)Kim, Kumar, Pavlovic, and Rowley}]{kim2008face}
\bibinfo{author}{M.~Kim}, \bibinfo{author}{S.~Kumar},
  \bibinfo{author}{V.~Pavlovic}, \bibinfo{author}{H.~Rowley},
\newblock \bibinfo{title}{Face tracking and recognition with visual constraints
  in real-world videos},
\newblock in: \bibinfo{booktitle}{{IEEE Conference on Computer Vision and
  Pattern Recognition (CVPR)}}, \bibinfo{organization}{IEEE},
  \bibinfo{year}{2008}, pp. \bibinfo{pages}{1--8}.
\bibitem[{Paragios(2003)}]{paragios2003level}
\bibinfo{author}{N.~Paragios},
\newblock \bibinfo{title}{A level set approach for shape-driven segmentation
  and tracking of the left ventricle},
\newblock \bibinfo{journal}{IEEE Transactions on Medical Imaging}
  \bibinfo{volume}{22} (\bibinfo{year}{2003}) \bibinfo{pages}{773--776}.
\bibitem[{P{\'e}rez et~al.(2002)P{\'e}rez, Hue, Vermaak, and
  Gangnet}]{perez2002color}
\bibinfo{author}{P.~P{\'e}rez}, \bibinfo{author}{C.~Hue},
  \bibinfo{author}{J.~Vermaak}, \bibinfo{author}{M.~Gangnet},
\newblock \bibinfo{title}{Color-based probabilistic tracking},
\newblock in: \bibinfo{booktitle}{{European Conference on Computer Vision
  (ECCV)}}, \bibinfo{publisher}{Springer}, \bibinfo{year}{2002}, pp.
  \bibinfo{pages}{661--675}.
\bibitem[{Nummiaro et~al.(2003)Nummiaro, Koller-Meier, and
  Van~Gool}]{nummiaro2003adaptive}
\bibinfo{author}{K.~Nummiaro}, \bibinfo{author}{E.~Koller-Meier},
  \bibinfo{author}{L.~Van~Gool},
\newblock \bibinfo{title}{An adaptive color-based particle filter},
\newblock \bibinfo{journal}{Image and vision computing} \bibinfo{volume}{21}
  (\bibinfo{year}{2003}) \bibinfo{pages}{99--110}.
\bibitem[{Wang et~al.(2013)Wang, Lu, and Yang}]{LSST}
\bibinfo{author}{D.~Wang}, \bibinfo{author}{H.~Lu}, \bibinfo{author}{M.-H.
  Yang},
\newblock \bibinfo{title}{Least soft-threshold squares tracking},
\newblock in: \bibinfo{booktitle}{{IEEE Conference on Computer Vision and
  Pattern Recognition (CVPR)}}, \bibinfo{organization}{IEEE},
  \bibinfo{year}{2013}, pp. \bibinfo{pages}{2371--2378}.
\bibitem[{Cehovin et~al.(2011)Cehovin, Kristan, and Leonardis}]{CoupleLayer}
\bibinfo{author}{L.~Cehovin}, \bibinfo{author}{M.~Kristan},
  \bibinfo{author}{A.~Leonardis},
\newblock \bibinfo{title}{An adaptive coupled-layer visual model for robust
  visual tracking},
\newblock in: \bibinfo{booktitle}{{International Conference on Computer Vision
  (ICCV)}}, \bibinfo{organization}{IEEE}, \bibinfo{year}{2011}, pp.
  \bibinfo{pages}{1363--1370}.
\bibitem[{Zhong et~al.(2014)Zhong, Lu, and Yang}]{zhong2014robust}
\bibinfo{author}{W.~Zhong}, \bibinfo{author}{H.~Lu}, \bibinfo{author}{M.-H.
  Yang},
\newblock \bibinfo{title}{{Robust Object Tracking via Sparse Collaborative
  Appearance Model}},
\newblock \bibinfo{journal}{{IEEE Transactions on Image Processing (TIP)}}
  (\bibinfo{year}{2014}) \bibinfo{pages}{2356--68}.
\bibitem[{Li et~al.(2007)Li, Ai, Yamashita, Lao, and Kawade}]{Li2007cascade}
\bibinfo{author}{Y.~Li}, \bibinfo{author}{H.~Ai},
  \bibinfo{author}{T.~Yamashita}, \bibinfo{author}{S.~Lao},
  \bibinfo{author}{M.~Kawade},
\newblock \bibinfo{title}{Tracking in low frame rate video: A cascade particle
  filter with discriminative observers of different lifespans},
\newblock in: \bibinfo{booktitle}{{IEEE Conference on Computer Vision and
  Pattern Recognition (CVPR)}}, \bibinfo{organization}{IEEE},
  \bibinfo{year}{2007}, pp. \bibinfo{pages}{1--8}.
\bibitem[{Jia et~al.(2012)Jia, Lu, and Yang}]{jia2012visual}
\bibinfo{author}{X.~Jia}, \bibinfo{author}{H.~Lu}, \bibinfo{author}{M.-H.
  Yang},
\newblock \bibinfo{title}{Visual tracking via adaptive structural local sparse
  appearance model},
\newblock in: \bibinfo{booktitle}{{IEEE Conference on Computer Vision and
  Pattern Recognition (CVPR)}}, \bibinfo{organization}{IEEE},
  \bibinfo{year}{2012}, pp. \bibinfo{pages}{1822--1829}.
\bibitem[{Dou and Li(2014)}]{dou2014robust}
\bibinfo{author}{J.~Dou}, \bibinfo{author}{J.~Li},
\newblock \bibinfo{title}{Robust visual tracking based on interactive multiple
  model particle filter by integrating multiple cues},
\newblock \bibinfo{journal}{Neurocomputing} \bibinfo{volume}{135}
  (\bibinfo{year}{2014}) \bibinfo{pages}{118--129}.
\bibitem[{Gilks(2005)}]{gilks2005markov}
\bibinfo{author}{W.~R. Gilks}, \bibinfo{title}{Markov chain monte carlo},
  \bibinfo{publisher}{Wiley Online Library}, \bibinfo{year}{2005}.
\bibitem[{Septier et~al.(2009)Septier, Pang, Carmi, and
  Godsill}]{MCMC-overview}
\bibinfo{author}{F.~Septier}, \bibinfo{author}{S.~K. Pang},
  \bibinfo{author}{A.~Carmi}, \bibinfo{author}{S.~Godsill},
\newblock \bibinfo{title}{On mcmc-based particle methods for bayesian
  filtering: Application to multitarget tracking},
\newblock in: \bibinfo{booktitle}{IEEE International Workshop on Computational
  Advances in Multi-Sensor Adaptive Processing (CAMSAP)},
  \bibinfo{organization}{IEEE}, \bibinfo{year}{2009}, pp.
  \bibinfo{pages}{360--363}.
\bibitem[{Khan et~al.(2004)Khan, Balch, and Dellaert}]{khan2004mcmc}
\bibinfo{author}{Z.~Khan}, \bibinfo{author}{T.~Balch},
  \bibinfo{author}{F.~Dellaert},
\newblock \bibinfo{title}{An mcmc-based particle filter for tracking multiple
  interacting targets},
\newblock in: \bibinfo{booktitle}{{European Conference on Computer Vision
  (ECCV)}}, \bibinfo{publisher}{Springer}, \bibinfo{year}{2004}, pp.
  \bibinfo{pages}{279--290}.
\bibitem[{Khan et~al.(2005)Khan, Balch, and Dellaert}]{khan2005mcmc}
\bibinfo{author}{Z.~Khan}, \bibinfo{author}{T.~Balch},
  \bibinfo{author}{F.~Dellaert},
\newblock \bibinfo{title}{Mcmc-based particle filtering for tracking a variable
  number of interacting targets},
\newblock \bibinfo{journal}{{IEEE Transactions on Pattern Recognition and
  Machine Intelligence (TPAMI)}} \bibinfo{volume}{27} (\bibinfo{year}{2005})
  \bibinfo{pages}{1805--1819}.
\bibitem[{Kwon and Lee(2013)}]{WLMC-PAMI}
\bibinfo{author}{J.~Kwon}, \bibinfo{author}{K.~M. Lee},
\newblock \bibinfo{title}{Wang-landau monte carlo-based tracking methods for
  abrupt motions},
\newblock \bibinfo{journal}{{IEEE Transactions on Pattern Recognition and
  Machine Intelligence (TPAMI)}} \bibinfo{volume}{35} (\bibinfo{year}{2013})
  \bibinfo{pages}{1011--1024}.
\bibitem[{Zhou et~al.(2012)Zhou, Lu, Lu, and Zhou}]{SAMC-TIP}
\bibinfo{author}{X.~Zhou}, \bibinfo{author}{Y.~Lu}, \bibinfo{author}{J.~Lu},
  \bibinfo{author}{J.~Zhou},
\newblock \bibinfo{title}{Abrupt motion tracking via intensively adaptive
  markov-chain monte carlo sampling},
\newblock \bibinfo{journal}{{IEEE Transactions on Image Processing (TIP)}}
  \bibinfo{volume}{21} (\bibinfo{year}{2012}) \bibinfo{pages}{789--801}.
\bibitem[{Zhou et~al.(2014)Zhou, Lu, and Di}]{zhou2014nearest}
\bibinfo{author}{T.~Zhou}, \bibinfo{author}{Y.~Lu}, \bibinfo{author}{H.~Di},
\newblock \bibinfo{title}{Nearest neighbor field driven stochastic sampling for
  abrupt motion tracking},
\newblock in: \bibinfo{booktitle}{{IEEE International Conference on Multimedia
  and Expo (ICME)}}, \bibinfo{organization}{IEEE}, \bibinfo{year}{2014}, pp.
  \bibinfo{pages}{1--6}.
\bibitem[{Yilmaz et~al.(2006)Yilmaz, Javed, and Shah}]{Survey-1}
\bibinfo{author}{A.~Yilmaz}, \bibinfo{author}{O.~Javed},
  \bibinfo{author}{M.~Shah},
\newblock \bibinfo{title}{Object tracking: A survey},
\newblock \bibinfo{journal}{Acm computing surveys (CSUR)} \bibinfo{volume}{38}
  (\bibinfo{year}{2006}) \bibinfo{pages}{13}.
\bibitem[{Yang et~al.(2011)Yang, Shao, Zheng, Wang, and Song}]{yang2011recent}
\bibinfo{author}{H.~Yang}, \bibinfo{author}{L.~Shao},
  \bibinfo{author}{F.~Zheng}, \bibinfo{author}{L.~Wang},
  \bibinfo{author}{Z.~Song},
\newblock \bibinfo{title}{Recent advances and trends in visual tracking: A
  review},
\newblock \bibinfo{journal}{Neurocomputing} \bibinfo{volume}{74}
  (\bibinfo{year}{2011}) \bibinfo{pages}{3823--3831}.
\bibitem[{Isard and Blake(1998)}]{isard1998condensation}
\bibinfo{author}{M.~Isard}, \bibinfo{author}{A.~Blake},
\newblock \bibinfo{title}{Condensation\textendash conditional density
  propagation for visual tracking},
\newblock \bibinfo{journal}{{International Journal on Computer Vision (IJCV)}}
  \bibinfo{volume}{29} (\bibinfo{year}{1998}) \bibinfo{pages}{5--28}.
\bibitem[{Philomin et~al.(2000)Philomin, Duraiswami, and
  Davis}]{philomin2000quasi}
\bibinfo{author}{V.~Philomin}, \bibinfo{author}{R.~Duraiswami},
  \bibinfo{author}{L.~S. Davis},
\newblock \bibinfo{title}{Quasi-random sampling for condensation},
\newblock in: \bibinfo{booktitle}{{European Conference on Computer Vision
  (ECCV)}}, \bibinfo{organization}{Springer}, \bibinfo{year}{2000}, pp.
  \bibinfo{pages}{134--149}.
\bibitem[{Su et~al.(2014)Su, Zhao, Zhao, and Gu}]{su2014abrupt}
\bibinfo{author}{Y.~Su}, \bibinfo{author}{Q.~Zhao}, \bibinfo{author}{L.~Zhao},
  \bibinfo{author}{D.~Gu},
\newblock \bibinfo{title}{Abrupt motion tracking using a visual saliency
  embedded particle filter},
\newblock \bibinfo{journal}{Pattern Recognition} \bibinfo{volume}{47}
  (\bibinfo{year}{2014}) \bibinfo{pages}{1826--1834}.
\bibitem[{Hua and Wu(2004)}]{hua2004multi}
\bibinfo{author}{G.~Hua}, \bibinfo{author}{Y.~Wu},
\newblock \bibinfo{title}{Multi-scale visual tracking by sequential belief
  propagation},
\newblock in: \bibinfo{booktitle}{{IEEE Conference on Computer Vision and
  Pattern Recognition (CVPR)}}, volume~\bibinfo{volume}{1},
  \bibinfo{organization}{IEEE}, \bibinfo{year}{2004}, pp.
  \bibinfo{pages}{I--826}.
\bibitem[{Sullivan et~al.(1999)Sullivan, Blake, Isard, and
  MacCormick}]{sullivan1999object}
\bibinfo{author}{J.~Sullivan}, \bibinfo{author}{A.~Blake},
  \bibinfo{author}{M.~Isard}, \bibinfo{author}{J.~MacCormick},
\newblock \bibinfo{title}{Object localization by bayesian correlation},
\newblock in: \bibinfo{booktitle}{{International Conference on Computer Vision
  (ICCV)}}, volume~\bibinfo{volume}{2}, \bibinfo{organization}{IEEE},
  \bibinfo{year}{1999}, pp. \bibinfo{pages}{1068--1075}.
\bibitem[{Barnes et~al.(2009)Barnes, Shechtman, Finkelstein, and
  Goldman}]{barnes2009patchmatch}
\bibinfo{author}{C.~Barnes}, \bibinfo{author}{E.~Shechtman},
  \bibinfo{author}{A.~Finkelstein}, \bibinfo{author}{D.~Goldman},
\newblock \bibinfo{title}{Patchmatch: a randomized correspondence algorithm for
  structural image editing},
\newblock \bibinfo{journal}{ACM Transactions on Graphics-TOG}
  \bibinfo{volume}{28} (\bibinfo{year}{2009}) \bibinfo{pages}{24}.
\bibitem[{Korman and Avidan(2011)}]{CSH}
\bibinfo{author}{S.~Korman}, \bibinfo{author}{S.~Avidan},
\newblock \bibinfo{title}{Coherency sensitive hashing},
\newblock in: \bibinfo{booktitle}{{International Conference on Computer Vision
  (ICCV)}}, \bibinfo{organization}{IEEE}, \bibinfo{year}{2011}, pp.
  \bibinfo{pages}{1607--1614}.
\bibitem[{He and Sun(2012)}]{he2012computing}
\bibinfo{author}{K.~He}, \bibinfo{author}{J.~Sun},
\newblock \bibinfo{title}{Computing nearest-neighbor fields via
  propagation-assisted kd-trees},
\newblock in: \bibinfo{booktitle}{{IEEE Conference on Computer Vision and
  Pattern Recognition (CVPR)}}, \bibinfo{organization}{IEEE},
  \bibinfo{year}{2012}, pp. \bibinfo{pages}{111--118}.
\bibitem[{Chen et~al.(2013)Chen, Cohen, Wu, Jin, and Lin}]{LDOF-NNF}
\bibinfo{author}{Z.~Chen}, \bibinfo{author}{S.~Cohen}, \bibinfo{author}{Y.~Wu},
  \bibinfo{author}{H.~Jin}, \bibinfo{author}{Z.~Lin},
\newblock \bibinfo{title}{Large displacement optical flow from nearest neighbor
  fields},
\newblock in: \bibinfo{booktitle}{{IEEE Conference on Computer Vision and
  Pattern Recognition (CVPR)}}, \bibinfo{organization}{IEEE},
  \bibinfo{year}{2013}, pp. \bibinfo{pages}{2443--2450}.
\bibitem[{Bao et~al.(2014)Bao, Yang, and Jin}]{bao2014edge}
\bibinfo{author}{L.~Bao}, \bibinfo{author}{Q.~Yang}, \bibinfo{author}{H.~Jin},
\newblock \bibinfo{title}{Fast edge-preserving patchmatch for large
  displacement optical flow},
\newblock \bibinfo{journal}{{IEEE Transactions on Image Processing (TIP)}}
  \bibinfo{volume}{23} (\bibinfo{year}{2014}) \bibinfo{pages}{4996--5006}.
\bibitem[{Hong et~al.(2013)Hong, Kwak, and Han}]{hong2013orderless}
\bibinfo{author}{S.~Hong}, \bibinfo{author}{S.~Kwak}, \bibinfo{author}{B.~Han},
\newblock \bibinfo{title}{Orderless tracking through model-averaged posterior
  estimation},
\newblock in: \bibinfo{booktitle}{{International Conference on Computer Vision
  (ICCV)}}, \bibinfo{organization}{IEEE}, \bibinfo{year}{2013}, pp.
  \bibinfo{pages}{2296--2303}.
\bibitem[{Wang and Lu(2012)}]{wang2012hamiltonian}
\bibinfo{author}{F.~Wang}, \bibinfo{author}{M.~Lu},
\newblock \bibinfo{title}{Hamiltonian monte carlo estimator for abrupt motion
  tracking},
\newblock in: \bibinfo{booktitle}{{IEEE International Conference on Pattern
  Recognition (ICPR)}}, \bibinfo{organization}{IEEE}, \bibinfo{year}{2012}, pp.
  \bibinfo{pages}{3066--3069}.
\bibitem[{Sundaram et~al.(2010)Sundaram, Brox, and Keutzer}]{Bro10e}
\bibinfo{author}{N.~Sundaram}, \bibinfo{author}{T.~Brox},
  \bibinfo{author}{K.~Keutzer},
\newblock \bibinfo{title}{Dense point trajectories by gpu-accelerated large
  displacement optical flow},
\newblock in: \bibinfo{booktitle}{{European Conference on Computer Vision
  (ECCV)}}, \bibinfo{publisher}{Springer}, \bibinfo{year}{2010}.
\bibitem[{Kristan et~al.(2011)Kristan, Leonardis, and
  Sko{\v{c}}aj}]{kristan2011multivariate}
\bibinfo{author}{M.~Kristan}, \bibinfo{author}{A.~Leonardis},
  \bibinfo{author}{D.~Sko{\v{c}}aj},
\newblock \bibinfo{title}{Multivariate online kernel density estimation with
  gaussian kernels},
\newblock \bibinfo{journal}{Pattern Recognition} \bibinfo{volume}{44}
  (\bibinfo{year}{2011}) \bibinfo{pages}{2630--2642}.
\bibitem[{Liang et~al.(2007)Liang, Liu, and Carroll}]{liang2007stochastic}
\bibinfo{author}{F.~Liang}, \bibinfo{author}{C.~Liu}, \bibinfo{author}{R.~J.
  Carroll},
\newblock \bibinfo{title}{Stochastic approximation in monte carlo computation},
\newblock \bibinfo{journal}{Journal of the American Statistical Association}
  \bibinfo{volume}{102} (\bibinfo{year}{2007}) \bibinfo{pages}{305--320}.
\bibitem[{Liang(2009)}]{liang2009improving}
\bibinfo{author}{F.~Liang},
\newblock \bibinfo{title}{Improving samc using smoothing methods: Theory and
  applications to bayesian model selection problems},
\newblock \bibinfo{journal}{The Annals of Statistics} \bibinfo{volume}{37}
  (\bibinfo{year}{2009}) \bibinfo{pages}{2626--2654}.
\bibitem[{Kwon and Lee(2010)}]{kwon2010visual}
\bibinfo{author}{J.~Kwon}, \bibinfo{author}{K.~M. Lee},
\newblock \bibinfo{title}{Visual tracking decomposition},
\newblock in: \bibinfo{booktitle}{{IEEE Conference on Computer Vision and
  Pattern Recognition (CVPR)}}, \bibinfo{organization}{IEEE},
  \bibinfo{year}{2010}, pp. \bibinfo{pages}{1269--1276}.
\bibitem[{Wu et~al.(2013)Wu, Lim, and Yang}]{wu2013online}
\bibinfo{author}{Y.~Wu}, \bibinfo{author}{J.~Lim}, \bibinfo{author}{M.-H.
  Yang},
\newblock \bibinfo{title}{Online object tracking: A benchmark},
\newblock in: \bibinfo{booktitle}{{IEEE Conference on Computer Vision and
  Pattern Recognition (CVPR)}}, \bibinfo{organization}{IEEE},
  \bibinfo{year}{2013}, pp. \bibinfo{pages}{2411--2418}.
\bibitem[{Babenko et~al.(2011)Babenko, Yang, and Belongie}]{babenko2011robust}
\bibinfo{author}{B.~Babenko}, \bibinfo{author}{M.-H. Yang},
  \bibinfo{author}{S.~Belongie},
\newblock \bibinfo{title}{Robust object tracking with online multiple instance
  learning},
\newblock \bibinfo{journal}{{IEEE Transactions on Pattern Recognition and
  Machine Intelligence (TPAMI)}} \bibinfo{volume}{33} (\bibinfo{year}{2011})
  \bibinfo{pages}{1619--1632}.

\end{thebibliography}
